\documentclass[sigconf, nonacm]{acmart}

\newcommand\vldbdoi{}

\newcommand\vldbavailabilityurl{https://olgaovcharenko.github.io/sempiper/}
\newcommand\vldbpagestyle{plain}

\usepackage{booktabs} 
\usepackage{dirtytalk}
\usepackage[utf8]{inputenc}
\usepackage{enumitem}
\usepackage{xcolor}
\usepackage{amsmath}
\usepackage{multirow} 
\usepackage{subfigure}
\usepackage{cleveref}
\usepackage{listings}
\usepackage{fancyvrb}
\usepackage{xurl}
\definecolor{darkgray}{rgb}{0.33, 0.33, 0.33}

\usepackage{balance}

\lstnewenvironment{Python}[1][]
  {\lstset{language=Python,
    basicstyle      = \scriptsize\ttfamily,
    keywordstyle    = \color{blue},
    keywordstyle    = [2] \color{teal}, 
    stringstyle     = \color{magenta},
    literate={ü}{{\"u}}1 {ö}{{\"o}}1 {É}{{\'E}}1 {œ}{{\oe}}1,
    deletekeywords=[2]{compile},    
    commentstyle    = \color{darkgray}\ttfamily,
           morekeywords=as,
           morekeywords=with,
           #1}%
  }
  {}

\usepackage{tikz}
\newcommand*\circled[1]{\protect\tikz[baseline=(char.base)]{\protect\node[shape=circle,fill=black,draw,inner sep=0.6pt] (char) {\textcolor{white}{\footnotesize \textbf{#1}}};}}

\newcommand{\header}[1]{\vspace{1mm}\noindent\textbf{#1}.}
\newcommand{\headerl}[1]{\vspace{1mm}\noindent\textit{#1}.}
\newcommand{\headerul}[1]{\vspace{1mm}\noindent\textit{\underline{#1}}.}

\begin{document}

\title{\textsc{SemPiper}: Interactive Code Synthesis for Semantic~Operators~in~Machine~Learning~Pipelines}
\author{Olga Ovcharenko}
\affiliation{%
  \institution{BIFOLD \& TU Berlin}
}
\email{ovcharenko@tu-berlin.de}

\author{Luciano Duarte}
\affiliation{%
  \institution{BIFOLD \& TU Berlin}
}
\email{duarte.castineira@tu-berlin.de}

\author{Sebastian Schelter}
\orcid{0000-0003-4722-5840}
\affiliation{%
  \institution{BIFOLD \& TU Berlin}
}
\email{schelter@tu-berlin.de}

\begin{abstract}


Machine learning (ML) pipelines require extensive data preparation, feature engineering, and integration across heterogeneous sources, making them tedious and error-prone to develop. While large language models (LLMs) have recently shown promise for assisting programming tasks, chat-based interfaces provide limited control over pipeline behavior and often produce code that is difficult to optimize or integrate into production systems.
We demonstrate \textsc{SemPipes}, a novel programming model that extends ML pipelines with declarative, LLM-powered semantic data operators. \textsc{SemPipes} allows developers to specify high-level natural language instructions for data-centric operations, while seamlessly combining these operators with arbitrary Python code from standard data science libraries. For the semantic operators, it synthesizes specialized implementations at pipeline training time, conditioned on dataset characteristics and pipeline context, enabling the flexible yet controlled integration of LLM capabilities.
We demonstrate \textsc{SemPipes} through \textsc{SemPiper}, an interactive interface that visualizes computational graphs of the pipelines, synthesized operator implementations, and optimization trajectories produced by an evolutionary search procedure. Attendees can explore three end-to-end scenarios, modify pipelines, inspect generated code, and observe how semantic operators are synthesized and iteratively optimized. The demonstration highlights how declarative semantic operators enable controllable, optimizable, and practical integration of LLMs into ML pipeline development.
\end{abstract}

\maketitle

\pagestyle{\vldbpagestyle}

\ifdefempty{\vldbavailabilityurl}{}{
\vspace{.3cm}
\begingroup\small\noindent\raggedright\textbf{Artifact Availability:}\\
The source code, data, and/or other artifacts have been made available at {\color{blue}\url{\vldbavailabilityurl}}.
\endgroup
}

\section{Introduction}
\label{sec:intro}

Large language models (LLMs) are reshaping software development through advances in code synthesis. Approaches range from AI-assistants for programming~\cite{anthropic2025claudecode} to fully autonomous program generation~\cite{chan2024mle,fang2025mlzero}. Despite their promise, these approaches raise concerns about control, safety, and unintended behavior~\cite{nolan2025replit}, and may even reduce developer productivity~\cite{becker2025measuringimpactearly2025ai}. A key limitation is that chat-based interaction only provides coarse control over the development process and produces code that is difficult to optimize.

\header{ML Pipelines} The outlined issues are particularly important for iterative, data-centric, and repetitive machine learning (ML) pipeline development, which is central to ML research and practice~\cite{van2024tabular}. In real-world settings, such pipelines operate on complex datasets originating from heterogeneous sources—including data lakes, warehouses, and log files, and must be joined, cleaned, augmented, and encoded with \emph{ML pipelines}~\cite{polyzotis2017data}. These pipelines encompass tasks such as prediction, integration, debugging, and postprocessing, combine complex logic across multiple libraries~\cite{psallidas2022data}, and are costly to execute in production~\cite{xin2021production}. Consequently, building tabular ML pipelines is tedious and error-prone, requiring substantial domain expertise and engineering effort. Industry surveys consistently identify data preparation as the most time-consuming and least enjoyable aspect of data science.

\header{\textsc{SemPipes}: Extending ML Pipelines with Semantic Data Operators} Recent work in semantic query processing points to a promising direction: integrating LLM-powered operators into data processing systems. In this paradigm, SQL is extended with semantic operators that invoke LLMs directly within database engines, enabling queries over multimodal and unstructured data~\cite{patel2024semantic,liu2025palimpzest,googlecloud_bigquery_generativeAI_overview_2025,snowflake_cortex_aisql_2025}. Inspired by this line of work, we recently introduced \textsc{SemPipes}~\cite{ovcharenko2026sempipes}, a novel programming model that extends tabular ML pipelines with declarative, LLM-powered \emph{semantic data operators} (\texttt{SemOps}). \textsc{SemPipes} pipelines are written in Python and selectively delegate data-centric operations to context-aware, optimizable semantic operators  that generate task-specific Python code (\Cref{sec:system}). Following a declarative paradigm, \textsc{SemPipes} separates \emph{what} should be computed from \emph{how} it is executed: developers provide high-level natural language instructions for selected operations, while the system synthesizes their concrete implementations.
During training, \textsc{SemPipes} uses LLMs to generate custom implementations for SemOps, conditioned on data characteristics, user instructions, and pipeline context. Therefore, \textsc{SemPipes} only requires a small number of LLM calls at training time, proportional to the number of semantic operators, and eliminates the reliance on LLMs during inference. This design enables semi-automated pipeline construction while preserving the ability to iteratively develop a pipeline in interactive notebook environments. Once defined, the operator implementations in a \textsc{SemPipes} pipeline can be further refined through LLM-based code mutation guided by evolutionary search to improve end-to-end predictive performance.
In contrast to systems like LOTUS~\cite{patel2024semantic} and Palimpzest~\cite{liu2025palimpzest}, which rely on LLM-based data processing, \textsc{SemPipes} leverages code generation.

\begin{figure}[t!]
  \centering
  \includegraphics[width=\columnwidth]{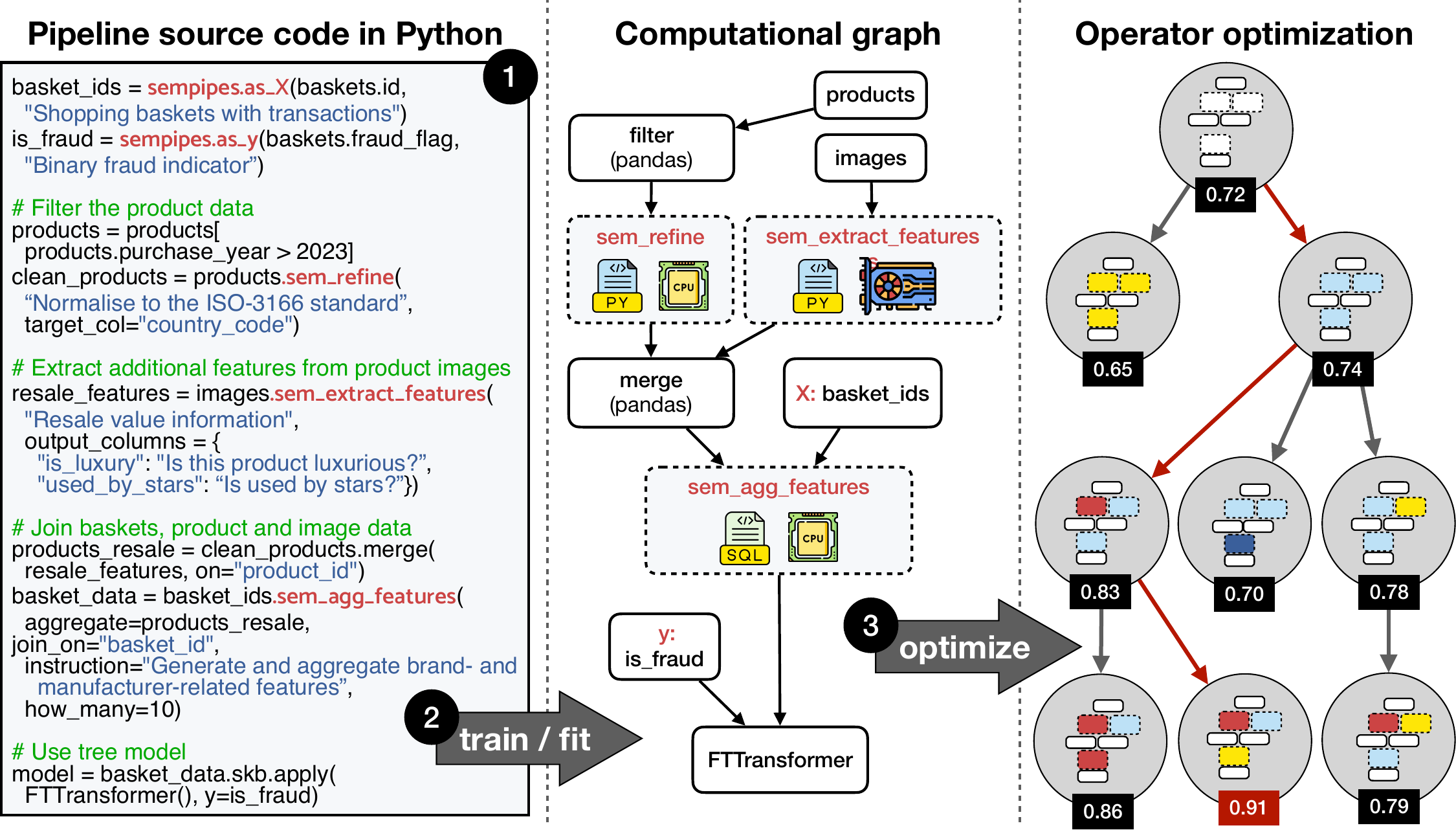}
  \vspace{-0.3cm}
  \caption{\textbf{High-level overview of \textsc{SemPipes}}. \textsc{SemPipes} extends standard Python ML pipelines with semantic data operators that delegate selected data-centric tasks to LLMs. During training, it synthesizes operator implementations from data, natural language instructions, and pipeline context, and can further optimize them on a validation set using evolutionary search guided by downstream model performance.}
  \label{fig:overview}
  \vspace{-0.8cm}  
\end{figure}

\header{Demonstration Details} We present \textsc{SemPiper}, an interactive demo that \say{\emph{plays} \textsc{SemPipes}} and exposes its internals through a dedicated user interface. We demonstrate \textsc{SemPipes} across three end-to-end ML scenarios with respect to fraud detection in e-commerce, integration of multimodal art data and prediction of house prices based on tables and images (\Cref{sec:demo}). Each scenario emphasizes different types of semantic data operators, ranging from semantic feature generation and extraction to data cleaning and refinement. For each scenario, we present the pipeline code, its corresponding computational graph, and the integrated SemOps. Through the interactive \textsc{SemPiper} interface, attendees can observe how operator implementations are synthesized at training time with \textsc{SemPipes}, inspect the generated Python code, and examine how each implementation is conditioned on dataset characteristics and natural language instructions. Attendees can further modify pipeline code, adjust instructions, or even implement their own scenario. In addition to interactive code synthesis for a pipeline, we also showcase the evolutionary optimizer for SemOps in \textsc{SemPipes}. For that, \textsc{SemPiper} visualizes different optimization trajectories, including a a search tree of candidate implementations, along with their validation performance and the evolutionary search procedure that selects improved variants. We provide the web-based user interface for our demonstration \textsc{SemPiper}, together with all example pipelines and datasets, at \textcolor{blue}{\url{https://olgaovcharenko.github.io/sempiper}}.
\section{System Overview}
\label{sec:system}

We briefly introduce \textsc{SemPipes}~\cite{ovcharenko2026sempipes}, available at \textcolor{blue}{\url{https://github.com/deem-data/sempipes}}.

\header{Overview} \Cref{fig:overview} provides a brief overview of \textsc{SemPipes}: \circled{1}~ML pipelines are defined in Python using standard data science libraries, including pandas for tabular data manipulation, NumPy for numerical computation, scikit-learn for feature encoding, and models compatible with the scikit-learn ecosystem. \textsc{SemPipes} extends these pipelines with {\em semantic data operators}, which {\em delegate selected data-centric operations to LLMs}. \circled{2}~During training, when a pipeline is fitted to the training data, \textsc{SemPipes} {\em synthesizes custom implementations for the semantic operators} based on data, prompt, and pipeline context. The synthesized code may execute locally on the CPU or leverage a pretrained model on a local GPU. \circled{3}~Once a pipeline is defined, {\em the implementations of its semantic operators can be tuned on a validation set}. \textsc{SemPipes} performs evolutionary search under a given search policy (e.g., Monte Carlo tree search), using downstream validation performance as the fitness function. During search, operator implementations are iteratively mutated via reflective prompting using prior performance scores and a tree-structured memory. 

\header{Implementation} \textsc{SemPipes} builds on skrub DataOps pipelines,\footnote{\url{https://skrub-data.org/stable/data_ops.html}}
 a lightweight abstraction for multi-table ML pipelines over dataframes. DataOps represents pipelines as a computational graph of data operations and dependencies, enabling lazy execution and pipeline rewriting. Unlike standard scikit-learn pipelines that are restricted to a linear sequence of single-table transformations, the DataOps abstraction elevates the entire pipeline, including multi-table dataframe operations, into a unified scikit-learn-style predictor with \texttt{fit} and \texttt{predict} methods. \textsc{SemPipes} integrates semantic data operators as scikit-learn-style estimators within the skrub computational graph.

\begin{table}[t!]
\centering
\caption{\textbf{Semantic data operators in \textsc{SemPipes}.}
}
\setlength{\tabcolsep}{6pt}   
\vspace{-3.5mm}
{\small
\begin{tabular}{ll}
\toprule
\multirow{1}{*}{\textbf{Category}} & \multirow{1}{*}{\textbf{Semantic Operators}}\\
\midrule
Data Cleaning & \texttt{sem\_fillna}, \texttt{sem\_clean}, \texttt{sem\_refine}\\
Feature Extraction & \texttt{sem\_extract\_features}, \texttt{sem\_gen\_features},\\
\& Generation & \texttt{sem\_agg\_features}\\
Data Augmentation & \texttt{sem\_augment}\\
Parameterization &  \texttt{sem\_select}, \texttt{sem\_choose}\\
\bottomrule
\end{tabular}}
\label{tab:semops}
\vspace{-0.5cm}
\end{table}

\header{Operator-Specific Code Synthesis at Training Time} 
\textsc{SemPipes} synthesizes operator implementations at training time, conditioned on input data characteristics, natural language instructions, and pipeline context, including the downstream task and model type. \Cref{tab:semops} shows the complete list of \texttt{SemOps}. 
The synthesized \texttt{SemOps} code varies by operator type and user instruction: for feature generation, \textsc{SemPipes} produces CPU-based pandas code; for multimodal feature extraction, \texttt{SemOps} leverage pretrained models on a local GPU, e.g., HuggingFace models in zero-shot mode. This code synthesis approach only requires a few LLM calls at training time, whose number is linear in the number of semantic operators. During inference,  the generated operator code is executed without any further LLM involvement. The operator-centric design enables fine-grained validation of synthesized code. Beyond syntax checking via parsing, \textsc{SemPipes} executes each operator on a data sample and verifies operator-specific output constraints, such as the expected shape of the resulting dataframe.

\header{Optimizing the \texttt{SemOps} in an ML Pipeline}  In practice, improving the data operations of a tabular ML pipeline is an iterative, manual process: data scientists select a target operation (e.g., feature generation), rewrite its code, potentially integrate external libraries, and rerun the pipeline on a validation set to assess performance improvements. \textsc{SemPipes} automates this workflow through LLM-based mutation of synthesized operator code, guided by tree-structured evolutionary search (e.g., Monte Carlo Tree Search). During this optimization, the system generates a set of candidate implementations by mutating the current code, integrates each candidate into the pipeline, and evaluates the generated pipeline variant on a validation set. The search procedure prioritizes promising code variants based on their impact on downstream model performance.
\section{Demonstration Details}
\label{sec:demo}

\begin{figure*}[t!]
  \centering
  \includegraphics[width=\textwidth]{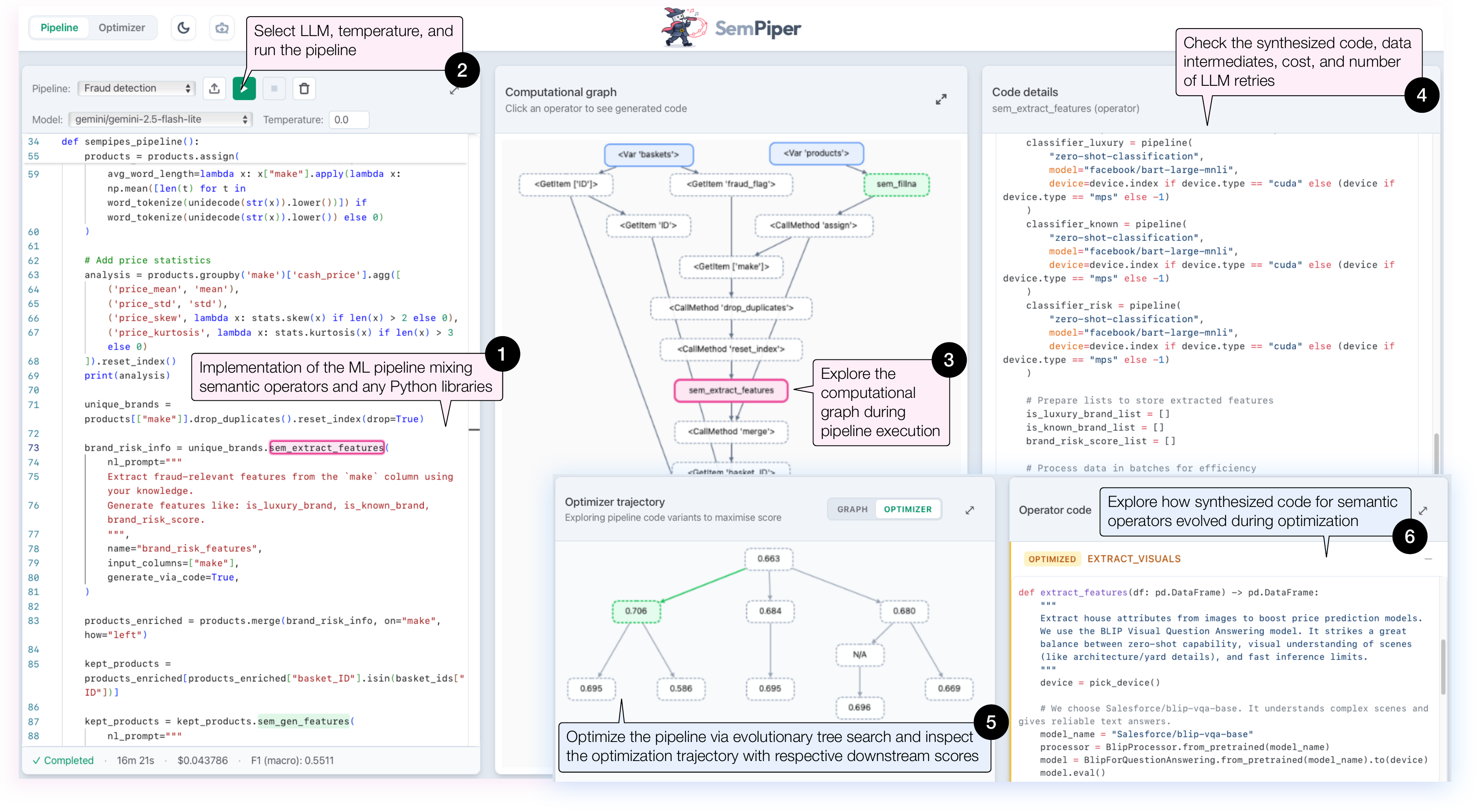}

  \vspace{-0.5cm}
  \caption{
  Web-based user interface of \text{SemPiper}: In the \textit{Pipeline} tab, \circled{1}~attendees load one of three provided pipelines or implement their own using any Python libraries. Next, \circled{2}~attendees select an LLM and temperature and execute the pipeline. \circled{3}~During execution, the computational graph of the pipeline can be  interactively explored. \circled{4}~Subsequently, attendees can deeply inspect the synthesized operator code, data intermediates, model performance, runtime, and cost of the pipeline. 
  In the \textit{Optimizer} tab, \circled{5}~attendees explore evolutionarily generated, evaluated pipeline variations, including their \circled{6}~their evolved operator code and performance across optimizer iterations.
  }
  \label{fig:app-both}

  \vspace{-0.3cm}
\end{figure*}

\textsc{SemPiper} demonstrates \textsc{SemPipes} through three scenarios, each centered on a ML pipeline highlighting certain semantic data operators. We provide the source code, datasets, and the web-based \textsc{SemPiper} interface at \textcolor{blue}{\url{https://github.com/OlgaOvcharenko/sempiper}}.
We demonstrate each scenario as detailed in Figure~\ref{fig:app-both}:

\begin{enumerate}[leftmargin=*]
    \item We introduce \circled{1} the scenario, the pipeline code, \circled{3} its corresponding computational graph, and semantic data operators.
    \item Attendees first \circled{2} execute the pipeline with \textsc{SemPipes}, to interactively synthesize its SemOps implementations by a selected LLM during training. Next, through the \textsc{SemPiper} interface, attendees inspect the \circled{4} generated Python code for each SemOp, and observe how it is conditioned on dataset characteristics and natural language instructions.
    \item Then, attendees can modify the natural language instructions of the SemOps and rewrite the pipeline code, to experiment with synthesizing new or improved operator implementations.
    \item Finally, \circled{5} we showcase the pipeline optimizer by visualizing multiple candidate implementations for each operator, along with their validation performance and the evolutionary search procedure that selects improved variants. \circled{6} Attendees can compare operator code, trace performance improvements, and inspect the cost, number of LLM calls, and time used for to run the whole pipeline and separate operators.
\end{enumerate}

\header{Interactive Code Synthesis} We detail our three scenarios.

\headerul{(a) Fraud Detection on Multi-Table Retail Data}
The first scenario is a fraud detection pipeline that trains a binary classifier to identify fraudulent shopping baskets. The pipeline uses two input tables: \texttt{baskets}, which contains fraud labels, and \texttt{products}, which describes the actual products purchased within each basket. The ML pipeline performs several preprocessing steps using popular Python libraries such as NLTK and unidecode for textual normalization and SciPy for price statistics. These steps are combined with semantic data operators, followed by a tree-based CatBoost classifier.

\headerl{Demonstrated Semantic Operators} This scenario highlights four semantic operators. First, \texttt{sem\_fillna} imputes missing product manufacturer values by inferring them from attributes such as the title and description.  Second, \texttt{sem\_extract\_features} extracts additional information from product brand names (\texttt{is\_luxury\_brand} and \texttt{is\_known\_brand}, and \texttt{brand\_risk\_score}) indicating the potential brand’s  association with fraud. Third, \texttt{sem\_gen\_features} synthesizes additional features. Finally, \texttt{sem\_agg\_features} aggregates the product-level data into basket-level features. The basket-level features are used to train the tree-based fraud detection model. 

\headerl{Exemplary Synthesized Code} With Google's \texttt{gemini-2.5-flash} as backing LLM, \textsc{SemPipes} synthesizes operator implementations in this pipeline as follows. For \texttt{sem\_fillna}, it automatically evaluates several candidate imputation strategies and selects a non-parametric \texttt{KNNImputer} the best-performing model. The generated extraction code for \texttt{sem\_extract\_features} is applies a zero-shot classification model \texttt{facebook/bart-large-mnli}, and a text generation model \texttt{google/flan-t5-base} that assigns a risk score.
Third, \texttt{sem\_gen\_features} calculates \texttt{avg\_brand\_cash\_price} and \texttt{brand\_frequency}. Finally, \texttt{sem\_agg\_features} computes aggregate features such as the total cash price per basket.

\headerul{(b)\,Cultural\,Origin\,Prediction\,for\,Multimodal\,Museum\,Catalog\,Data} 

\noindent The second scenario demonstrates a museum artifact analytics pipeline that combines tabular data with unstructured textual descriptions to label the artworks from the Metropolitan Museum of Art with their cultural origin.  The pipeline operates on the \texttt{artworks} table containing attributes describing a piece of art, together with free-text descriptions. We first enrich the table with lightweight natural language processing features extracted using the spaCy library, including named entities (people, locations, cultural groups, and dates), noun phrases, and adjective density. Next, the pipeline applies several semantic data operators to transform the noisy museum metadata into structured features. In the end, the data is vectorized and used to train an FT-Transformer classifier, an adaptation of the Transformer architecture for tabular data.

\headerl{Demonstrated Semantic Operators} This scenario evolves around three semantic operators. First, \texttt{sem\_extract\_features} converts heterogeneous date strings into a structured temporal representation. Second, \texttt{sem\_gen\_features} enriches the dataset with additional predictive features.  Finally, \texttt{sem\_refine} standardizes the raw museum object names to address inconsistencies and high-cardinality values to produce cleaner categorical representations. 

\headerl{Exemplary Synthesized Code} With Google's \texttt{gemini-2.5-flash} as backing LLM, \textsc{SemPipes} synthesizes operator code as follows. First, \texttt{sem\_extract\_features} extracts the date values \texttt{year\_start}, \texttt{start\_is\_bce}, \texttt{year\_end}, and \texttt{end\_is\_bce} by combining a  HuggingFace model \texttt{google/flan-t5-small} with rule-based parsing. The synthesized code prompts the model to extract structured information from the raw date string, producing JSON fields such as century and start/end years. These outputs are then refined using regular expressions and deterministic rules to enforce precise interval conventions for centuries and BCE/CE ranges.
Second, \texttt{sem\_gen\_features} removes irrelevant columns and creates new features from existing attributes, including text length features, temporal features derived about artwork duration, and features such as \texttt{has\_country\_info} and \texttt{has\_artist\_nationality\_info}.
Finally, \texttt{sem\_refine} standardizes the raw museum object names using regular expressions and correspondence dictionaries.

\headerul{(c) Price Prediction on Multimodal Real Estate Data}
The third scenario demonstrates a house price prediction pipeline that combines tabular and image data to estimate the price of houses in California. The pipeline consumes three input tables: \texttt{facts}, containing house-level attributes such as price, square footage, and room counts; \texttt{cities}, containing location information; and \texttt{images}, containing image metadata and file paths. The scenario is a regression task, predicting the house selling price.
After joining the tables, the pipeline performs several manual preprocessing and feature engineering steps. The enriched dataset is then vectorized using the \texttt{TableVectorizer} from the skrub library and the consumed by the tabular foundation model \texttt{TabPFNRegressor}. Finally, predicted prices are combined with the original data and analyzed using DuckDB to compute city-level statistics.

\headerl{Demonstrated Semantic Operators} This scenario highlights three semantic operators. First, \texttt{sem\_clean} is instructed to detect and correct anomalies in the \texttt{sqft} column. Second, \texttt{sem\_extract\_features} derives features from house images, with respect to the house exterior, leveraging domain knowledge about housing characteristics and geographic context in California. Finally, \texttt{sem\_gen\_features} synthesizes additional predictive attributes by generating property-, city-, and location-level features.

\headerl{Exemplary Synthesized Code} With Google's \texttt{gemini-2.5-flash} as backing LLM, \textsc{SemPipes} synthesizes operator implementations in this pipeline as follows. The code for \texttt{sem\_clean} cleans the \texttt{sqft} column by removing outliers through IQR-based capping. The synthesized implementation of the \texttt{sem\_extract\_features} operator applies the popular visual question answering model \texttt{Salesforce/blip-vqa-base} to extract image-based  attributes. Finally, \texttt{sem\_gen\_features} synthesize four features (\texttt{total\_rooms}, \texttt{sqft\_per\_bath}, \texttt{city\_avg\_sqft}, and \texttt{city\_avg\_bed}), which capture both property-level characteristics and local housing context.

\header{Evolutionary Optimization of Semantic Operator Code} The final part of the demonstration centers on an interactive optimizer interface for exploring \textsc{SemPipes}' evolutionary tree search runs. The interface visualizes the search process under different strategies, such as Monte Carlo Tree Search, showing how candidate operator implementations are generated and evaluated over time. Each node in the search tree exposes the synthesized code and its predictive performance on the validation set, allowing attendees to trace how the optimizer navigates the search space and discovers on best-performing pipeline variants.
For example, for a house price prediction task, attendees can observe how the optimizer leverages the \texttt{Salesforce/blip-vqa-base} visual question answering model and refines which features to extract. The best-performing pipeline extracts semantically rich features such as \texttt{number\_of\_stories}, \texttt{landscape\_quality}, \texttt{architectural\_style}, \texttt{garage\_capacity}, and \texttt{driveway\_size}, whereas weaker variants relied on coarser features like \texttt{has\_visible\_garage} and \texttt{exterior\_condition}.





\end{document}